%% file: main-arxiv.tex
\documentclass[11pt]{article}
\usepackage[pdftex,hyperindex,bookmarks=false,draft,hidelinks,breaklinks]{hyperref}
\usepackage{acl2015}
\pdfoutput=1

\usepackage{times}
\usepackage{latexsym}
\usepackage{times}
\usepackage{helvet}
\usepackage{courier}
\usepackage{color}
\usepackage{comment}
\usepackage{booktabs}
\usepackage{graphicx}
\usepackage{verbatim}
\usepackage[mode=buildnew]{standalone}
\usepackage{amsmath}
\usepackage{multirow}
\usepackage{tikz}
\usepackage{mystuff-nlp}
\usepackage{subcaption}

\usetikzlibrary{calc}
\usetikzlibrary{positioning}

\renewcommand{\vec}[1]{\mathbf{#1}}
\newcommand{\example}[1]{\emph{#1}}
\newcommand{\thetitle}[0]{Confounds and Consequences in Geotagged Twitter Data}
\newcommand{\uma}[1]{[\textcolor{blue}{#1 -U}]}
\newcommand{\gpsmsa}{\textsc{GPS-MSA-balanced}}
\newcommand{\gpscty}{\textsc{GPS-County-balanced}}
\newcommand{\locmsa}{\textsc{Loc-MSA-balanced}}

\title{\thetitle}

\author{Umashanthi Pavalanathan \and Jacob Eisenstein \\
  School of Interactive Computing \\
  Georgia Institute of Technology \\
  Atlanta, GA 30308 \\
  {\tt \{umashanthi + jacobe\}@gatech.edu}}

\date{}

\begin{document}
\maketitle

\input{abstract}

\input{intro}
\input{data}

\input{biases}

\input{language}
\input{prediction}
\input{related}

\input{discussion}
\input{ack}

\bibliographystyle{acl}
\bibliography{cite-strings,cites,cite-definitions}
\end{document}

%% file: abstract.tex
\begin{abstract}
Twitter is often used in quantitative studies that identify geographically-preferred topics, writing styles, and entities. These studies rely on either GPS coordinates attached to individual messages, or on the user-supplied location field in each profile.
In this paper, we compare these data acquisition techniques and quantify the biases that they introduce; we also measure their effects on linguistic analysis and text-based geolocation. GPS-tagging and self-reported locations yield measurably different corpora, and these linguistic differences are partially attributable to differences in dataset composition by age and gender. Using a latent variable model to induce age and gender, we show how these demographic variables interact with geography to affect language use. We also show that the accuracy of text-based geolocation varies with population demographics, giving the best results for men above the age of 40.
\end{abstract}

%% file: intro.tex
\section{Introduction}
Social media data such as Twitter is frequently used to identify the unique characteristics of geographical regions, including topics of interest~\cite{hong2012discovering}, linguistic styles and dialects~\cite{eisenstein2010latent,gonccalves2014crowdsourcing}, political opinions~\cite{caldarelli2014multi}, and public health~\cite{broniatowski2013national}. Social media permits the aggregation of datasets that are orders of magnitude larger than could be assembled via traditional survey techniques, enabling analysis that is simultaneously fine-grained and global in scale. Yet social media is not a representative sample of any ``real world'' population, aside from social media itself. Using social media as a sample therefore risks introducing both geographic and demographic biases~\cite{mislove2011understanding,hecht2014tale,longley2015,malik_bias_2015}.

This paper examines the effects of these biases on the geo-linguistic inferences that can be drawn from Twitter. We focus on the ten largest metropolitan areas in the United States, and consider three sampling techniques: drawing an equal number of GPS-tagged tweets from each area; drawing a \emph{county-balanced} sample of GPS-tagged messages to correct Twitter's urban skew~\cite{hecht2014tale}; and drawing a sample of \emph{location-annotated} messages, using the location field in the user profile. Leveraging self-reported first names and census statistics, we show that the age and gender composition of these datasets differ significantly.

Next, we apply standard methods from the literature to identify geo-linguistic differences, and test how the outcomes of these methods depend on the sampling technique and on the underlying demographics. We also test the accuracy of text-based geolocation~\cite{cheng2010you,eisenstein2010latent} in each dataset, to determine whether the accuracies reported in recent work will generalize to more balanced samples.

The paper reports several new findings about geotagged Twitter data:
\begin{itemize}
\setlength\parskip{0pt}
\setlength\parsep{0pt}
\setlength\itemsep{1pt}
\item In comparison with tweets with self-reported locations, GPS-tagged tweets are written more often by young people and by women.
\item There are corresponding linguistic differences between these datasets, with GPS-tagged tweets including more geographically-specific non-standard words. 
\item Young people use significantly more geographically-specific non-standard words. Men tend to mention more geographically-specific entities than women, but these differences are significant only for individuals at the age of 30 or older.
\item Users who GPS-tag their tweets tend to write more, making them easier to geolocate. Evaluating text-based geolocation on GPS-tagged tweets probably overestimates its accuracy.
\item Text-based geolocation is significantly more accurate for men and for older people.
\end{itemize}
These findings should inform future attempts to generalize from geotagged Twitter data, and may suggest investigations into the demographic properties of other social media sites. 

We first describe the basic data collection principles that hold throughout the paper (\autoref{sec:data}). The following three sections tackle demographic biases (\autoref{sec:biases}), their linguistic consequences (\autoref{sec:ling}), and the impact on text-based geolocation (\autoref{sec:prediction}); each of these sections begins with a discussion of methods, and then presents results. We then summarize related work and conclude.

%% file: data.tex
\section{Dataset}
\label{sec:data}
This study is performed on a dataset of tweets gathered from Twitter's streaming API from February 2014 to January 2015. During an initial filtering step we removed retweets, repetitions of previously posted messages which contain the ``retweeted\_status'' metadata or ``RT'' token which is widely used among Twitter users to indicate a retweet. To eliminate spam and automated accounts~\cite{yardi2009detecting}, we removed tweets containing URLs, user accounts with more than 1000 followers or followees, accounts which have tweeted more than 5000 messages at the time of data collection, and the top 10\% of accounts based on number of messages in our dataset. We also removed users who have written more than 10\% of their tweets in any language other than English, using Twitter's \texttt{lang} metadata field. Exploration of code-switching~\cite{solorio2008learning} and the role of second-language English speakers~\cite{eleta2014multilingual} is left for future work.

We consider the ten largest Metropolitan Statistical Areas (MSAs) in the United States, listed in Table~\ref{tab:l1-distance}. MSAs are defined by the U.S. Census Bureau as geographical regions of high population with density organized around a single urban core; they are not legal administrative divisions. MSAs include outlying areas that may be substantially less urban than the core itself. For example, the Atlanta MSA is centered on Fulton County (1750 people per square mile), but extends to Haralson County (100 people per square mile), on the border of Alabama. A per-county analysis of this data therefore enables us to assess the degree to which Twitter's skew towards urban areas biases geo-linguistic analysis.

%% file: biases.tex
\section{Representativeness of geotagged Twitter data}
\label{sec:biases}
We first assess potential biases in sampling techniques for obtaining geotagged Twitter data. In particular, we compare two possible techniques for obtaining data: the location field in the user profile~\cite{poblete2011all,dredze2013carmen}, and the GPS coordinates attached to each message~\cite{cheng2010you,eisenstein2010latent}.

\newcommand{\gpsset}{$\mathcal{D}_G$}
\newcommand{\locset}{$\mathcal{D}_L$}

\subsection{Methods}
\begin{figure*}
\centering
\includegraphics[width=.96\textwidth]{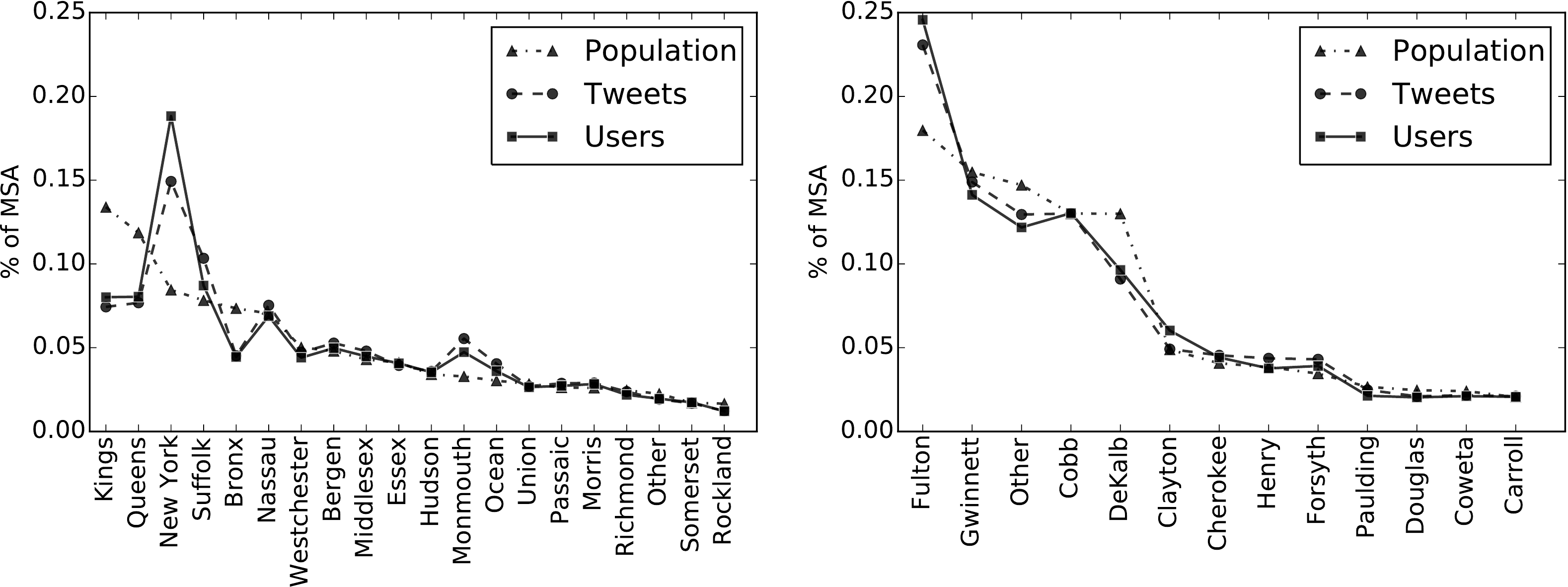}
\caption{Proportion of census population, Twitter messages, and Twitter user accounts, by county. New York is shown on the left, Atlanta on the right.}
\label{fig:nyc_pop}
\end{figure*}

To build a dataset of GPS-tagged messages, we extracted the GPS latitude and longitude coordinates reported in the tweet, and used \textsc{gis-tools}\footnote{{\url{https://github.com/DrSkippy/Data-Science-45min-Intros/blob/master/gis-tools-101/gis_tools.ipynb}}}
reverse geocoding to identify the corresponding counties.
This set of geotagged messages will be denoted \gpsset. Only 1.24\% of messages contain geo-coordinates, and it is possible that the individuals willing to share their GPS comprise a skewed population. We therefore also considered the user-reported location field in the Twitter profile, focusing on the two most widely-used patterns: (1) city name, (2) city name and two letter state name (e.g. \example{Chicago} and \example{Chicago, IL}). Messages that matched any of the ten largest MSAs were grouped into a second set, \locset. 

While the inconsistencies of writing style in the Twitter location field are well-known~\cite{hecht2011tweets}, analysis of the intersection between \gpsset~and \locset~found that the two data sources agreed the overwhelming majority of the time, suggesting that most self-provided locations are accurate. Of course, there may be many false negatives --- profiles that we fail to geolocate due to the use of non-standard toponyms like \example{Pixburgh} and \example{ATL}. If so, this would introduce a bias in the population sample in \locset. Such a bias might have linguistic consequences, with datasets based on the location field containing less non-standard language overall.

\subsubsection{Subsampling}
The initial samples \gpsset~and \locset~were then resampled to create the following balanced datasets:
\begin{description}
\setlength\itemsep{0pt}
\item[\gpsmsa] From \gpsset, we randomly sampled 25,000 tweets per MSA as the message-balanced sample, and all the tweets from  2,500 users per MSA as the user-balanced sample. Balancing across MSAs ensures that the largest MSAs do not dominate the linguistic analysis.
\item[\gpscty] We resampled \gpsset~based on county-level population (obtained from the U.S. Census Bureau), and again obtained message-balanced and user-balanced samples. These samples are more geographically representative of the overall population distribution across each MSA.
\item[\locmsa] From \locset, we randomly sampled 25,000 tweets per MSA as the message-balanced sample, and all the tweets from 2,500 users per MSA as the user-balanced sample. It is not possible to obtain county-level geolocations in \locset, as exact geographical coordinates are unavailable.
\end{description}

\subsubsection{Age and gender identification}
\label{sssec:age-prior}
To estimate the distribution of ages and genders in each sample, we queried statistics from the Social Security Administration, which records the number of individuals born each year with each given name. Using this information, we obtained the probability distribution of age values for each given name. We then matched the names against the first token in the name field of each user's profile, enabling us to induce approximate distributions over ages and genders. Unlike Facebook and Google+, Twitter does not have a ``real name'' policy, so users are free to give names that are fake, humorous, etc.
We eliminate user accounts whose names are not sufficiently common in the social security database (i.e.\ first names which are at least 100 times more frequent in Twitter than in the social security database), thereby omitting 33\% of user accounts, and 34\% of tweets.
While some individuals will choose names not typically associated with their gender, we assume that this will happen with roughly equal probability in both directions. So, with these caveats in mind, we induce the age distribution for the \textsc{GPS-MSA-Balanced} sample and the \textsc{Loc-MSA-Balanced} sample as,

\begin{small}
\begin{align}
p(a \mid \text{name} = n) & = \frac{count(\text{name} = n, \text{age} = a)}{\sum_{a'} count(\text{name} = n, \text{age} = a')} \\
p_{\mathcal{D}}(a) & \propto \sum_{i \in \mathcal{D}}p(a \mid \text{name} = n_i).
\end{align}
\end{small}

We induce distributions over author gender in much the same way~\cite{mislove2011understanding}. This method does not incorporate prior information about the ages of Twitter users, and thus assigns too much probability to the extremely young and old, who are unlikely to use the service. While it would be easy to design such a prior --- for example, assigning zero prior probability to users under the age of five or above the age of 95 --- we see no principled basis for determining these cutoffs. We therefore focus on the \emph{differences} between the estimated $p_{\mathcal{D}}(a)$ for each sample $\mathcal{D}$. 

\subsection{Results}

\paragraph{Geographical biases in the GPS Sample} 
We first assess the differences between the true population distributions over counties, and the per-tweet and per-user distributions. Because counties vary widely in their degree of urbanization and other demographic characteristics, this measure is a proxy for the representativeness of GPS-based Twitter samples (county information is not available for the \locmsa~sample). 
Population distributions for New York and Atlanta are shown in~\autoref{fig:nyc_pop}. In Atlanta, Fulton County is the most populous and most urban, and is overrepresented in both geotagged tweets and user accounts; most of the remaining counties are correspondingly underrepresented. This coheres with the urban bias noted earlier by \newcite{hecht2014tale}. In New York, Kings County (Brooklyn) is the most populous, but is underrepresented in both the number of geotagged tweets and user accounts, at the expense of New York County (Manhattan). Manhattan is the commercial and entertainment center of the New York MSA, so residents of outlying counties may be tweeting from their jobs or social activities.

\begin{table}
  \small
  \begin{center}
    {
    \scalebox{1}{
		\begin{tabular}{llll} \toprule
		 & \textbf{Num.}& \textbf{L1 Dist.}&\textbf{L1 Dist.}\\
		\textbf{MSA}& \textbf{Counties}&\textbf{Population}&\textbf{Population}\\
		& &\textbf{vs. Users}&\textbf{vs. Tweets}\\
		
		\midrule
		New York & 	23 	& 0.2891 	& 0.2825\\
		Los Angeles	 &	2	& 0.0203	& 0.0223\\
		Chicago	 &	14	& 0.0482	& 0.0535\\
		Dallas	 &	12	& 0.1437	& 0.1176\\
		Houston	 &	10	& 0.0394	& 0.0472\\
		Philadelphia	 &	11	& 0.1426	& 0.1202\\
		Washington DC	 &	22	& 0.2089	& 0.2750\\
		Miami	 &	3	& 0.0428	& 0.0362\\
		Atlanta	 &	28	& 0.1448	& 0.1730\\
		Boston	 &	7	& 0.1878	& 0.2303\\		
		\bottomrule
		\end{tabular}
		}
        \vspace{-5pt}
    \caption{L1 distance between county-level population and Twitter users and messages}\label{tab:l1-distance}
	}
        \vspace{-12pt}
  \end{center}
  \end{table}

To quantify the representativeness of each sample, we use the L1 distance $||\vec{x} - \vec{y}||_1 = \sum_c |p_c - t_c|$, where $p_c$ is the proportion of the MSA population residing in county $c$ and $t_c$ is the proportion of tweets (\autoref{tab:l1-distance}). County boundaries are determined by states, and their density varies: for example, the Los Angeles MSA covers only two counties, while the smaller Atlanta MSA is spread over 28 counties. The table shows that while New York is the most extreme example, most MSAs feature an asymmetry between county population and Twitter adoption.

\paragraph{Usage}
\begin{figure}
\includegraphics[width=.5\textwidth]{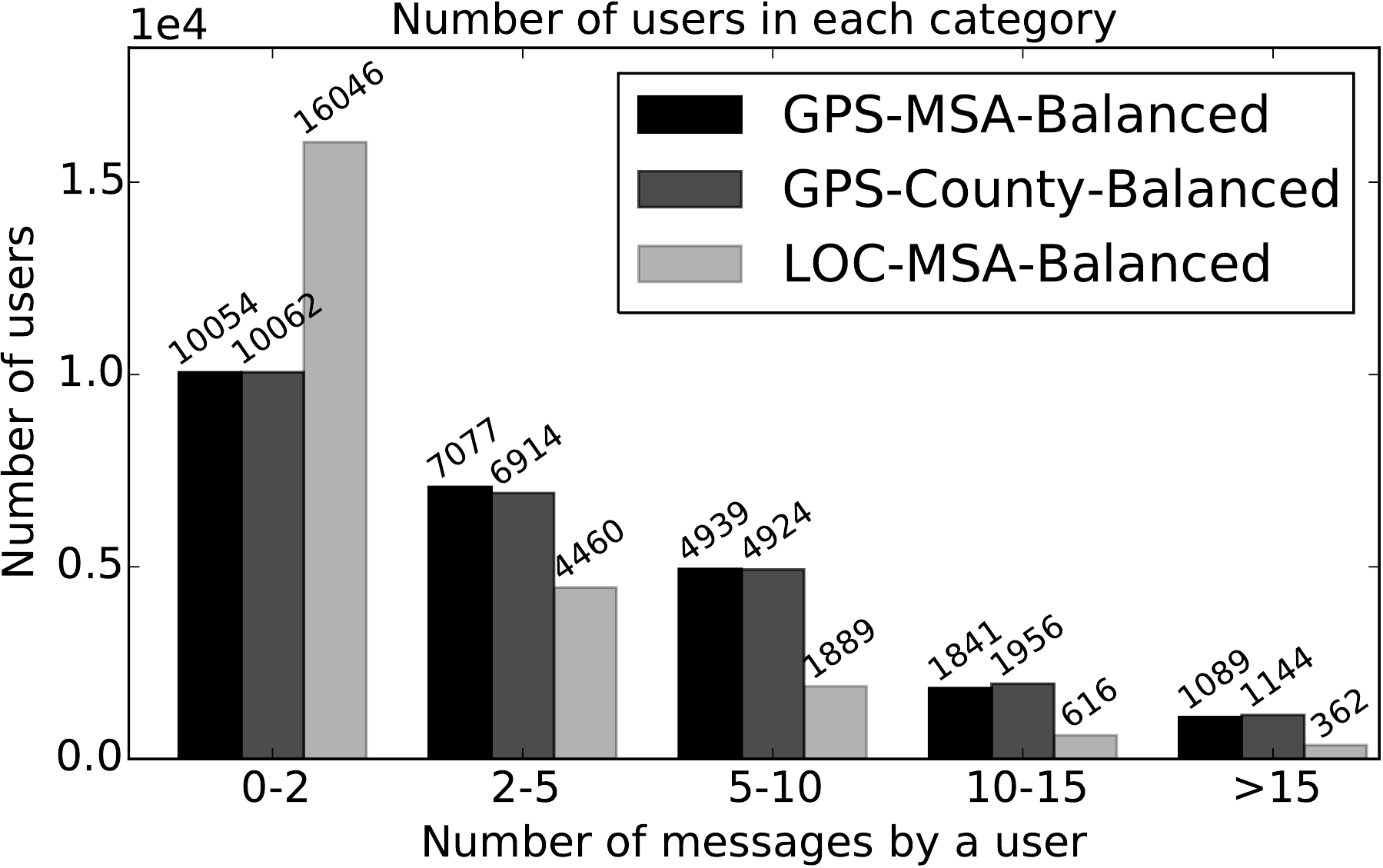}
\caption{User counts by number of Twitter messages}
\label{fig:usage}
\end{figure}

Next, we turn to differences between the GPS-based and profile-based techniques for obtaining ground truth data. As shown in \autoref{fig:usage}, the \locmsa\ sample contains more low-volume users than either the \gpsmsa\ or \gpscty\ samples. We can therefore conclude that the county-level geographical bias in the GPS-based data does not impact usage rate, but that the difference between GPS-based and profile-based sampling does; the linguistic consequences of this difference will be explored in the following sections.

\paragraph{Demographics} 
\begin{table*}
  \begin{center}
    {
		\begin{tabular}{lllll} \toprule
		\textbf{Sample}& \textbf{Expected Age}&\textbf{95\% CI} &\textbf{\% Female} & \textbf{95\% CI} \\
		\midrule
		\gpsmsa & 	36.17	& [36.07 -- 36.27] & 51.5 & [51.3 -- 51.8]\\
		\gpscty & 	36.25	& [36.16 -- 36.30] & 51.3 & [51.1 -- 51.6]\\
		\locmsa	 &	38.35	& [38.25 -- 38.44] & 49.3 & [49.1 -- 49.6]\\
		\bottomrule
		\end{tabular}
}		
    \caption{Demographic statistics for each dataset}
    \vspace{-8pt}
    \label{tab:dataset-demo}
 \end{center}
 \end{table*}

\begin{figure}
\includegraphics[width=.5\textwidth]{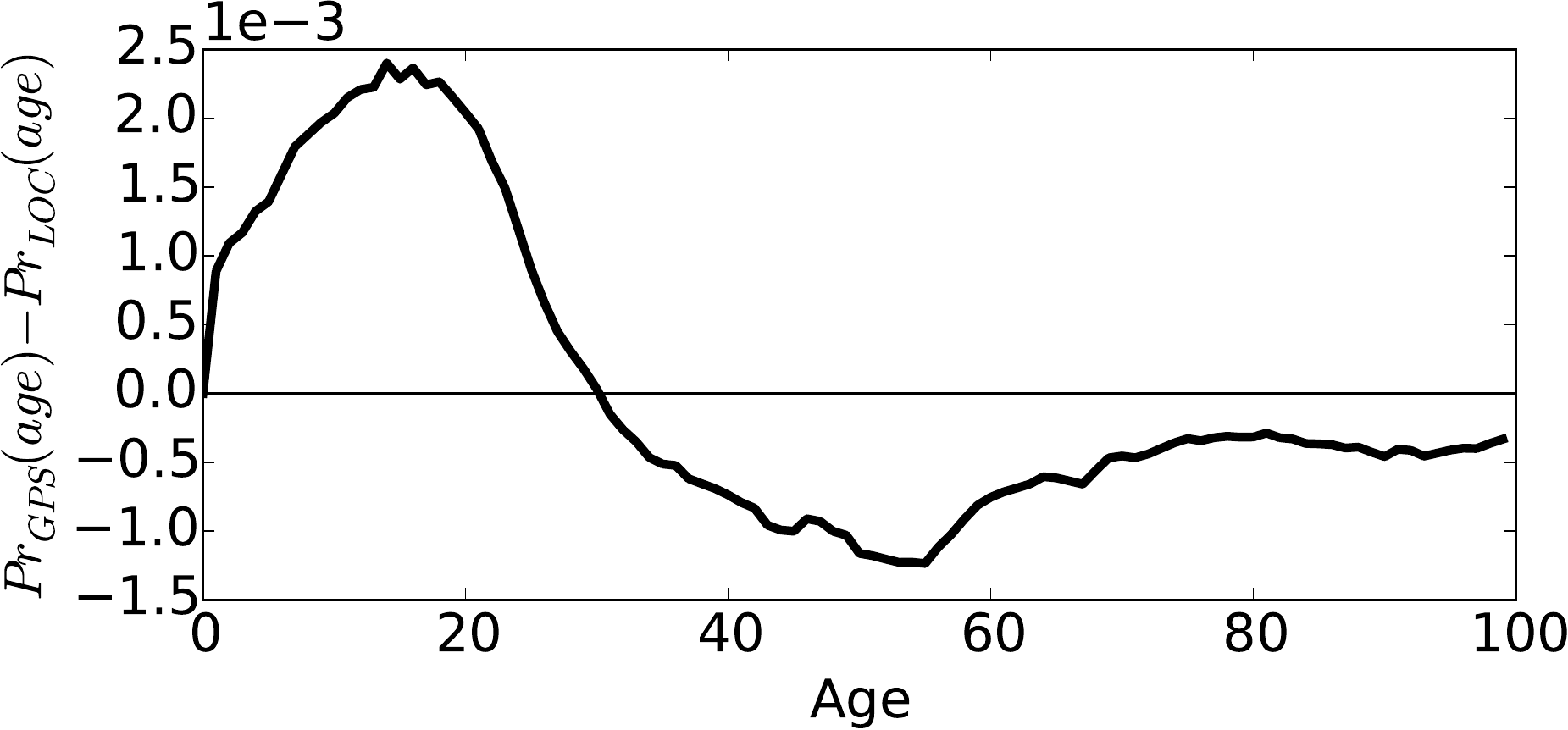}
\caption{Difference in age probability distributions between \gpsmsa\ and \locmsa.} 
\label{fig:age-diff}
\vspace{-6pt}
\end{figure}

Table~\ref{tab:dataset-demo} shows the expected age and gender for each dataset, with bootstrap confidence intervals. Users in the \locmsa\ dataset are on average two years older than in the \gpsmsa\ and \gpscty\ datasets, which are statistically indistinguishable. Focusing on the difference between \gpsmsa\ and \locmsa, we plot the difference in age probabilities in Figure~\ref{fig:age-diff}, showing that \gpsmsa\ includes many more teens and people in their early twenties, while \locmsa\  includes more people at middle age and older. Young people are especially likely to use social media on cellphones~\cite{lenhart2015mobile}, where location tagging would be more relevant than when Twitter is accessed via a personal computer. Social media users in the age brackets 18-29 and 30-49 are also more likely to tag their locations in social media posts than social media users in the age brackets 50-64 and 65+~\cite{zickuhr2013location}, with women and men tagging at roughly equal rates. Table~\ref{tab:dataset-demo} shows that the \gpsmsa\ and \gpscty\  samples contain significantly more women than \locmsa, though all three samples are close to 50\%.

%% file: language.tex
\section{Impact on linguistic generalizations}
\label{sec:ling}
Many papers use Twitter data to draw conclusions about the relationship between language and geography. What role do the demographic differences identified in the previous section have on the linguistic conclusions that emerge? We measure the differences between the linguistic corpora obtained by each data acquisition approach. Since the \gpsmsa\ and \gpscty\ methods have nearly identical patterns of usage and demographics, we focus on the difference between \gpsmsa\ and \locmsa. These datasets differ in age and gender, so we also directly measure the impact of these demographic factors on the use of geographically-specific linguistic variables.

\subsection{Methods}
\paragraph{Discovering geographical linguistic variables}
We focus on lexical variation, which is relatively easy to identify in text corpora. \newcite{monroe2008fightin} survey a range of alternative statistics for finding lexical variables, demonstrating that a regularized log-odds ratio strikes a good balance between distinctiveness and robustness. A similar approach is implemented in SAGE~\cite{eisenstein2011sparse}\footnote{https://github.com/jacobeisenstein/jos-gender-2014}, which we use here. For each sample --- \gpsmsa\ and \locmsa\ --- we apply SAGE to identify the twenty-five most salient lexical items for each metropolitan area.

\paragraph{Keyword annotation}
Previous research has identified two main types of geographical lexical variables. The first are non-standard words and spellings, such as \example{hella} and \example{yinz}, which have been found to be very frequent in social media~\cite{eisenstein2015geographical}. Other researchers have focused on the ``long tail'' of entity names~\cite{roller2012supervised}. A key question is the relative importance of these two variable types, since this would decide whether geo-linguistic differences are primarily topic-based or stylistic. It is therefore important to know whether the frequency of these two variable types depends on properties of the sample. To test this, we take the lexical items identified by SAGE (25 per MSA, for both the \gpsmsa\ and \locmsa\ samples), and annotate them as \annot{Nonstandard-Word}, \annot{Entity-Name}, or \annot{Other}. Annotation for ambiguous cases is based on the majority sense in randomly-selected examples. Overall, we identify 24 \annot{Nonstandard-Word}s and 185 \annot{Entity-Name}s.

\begin{figure}
  \begin{subfigure}[b]{0.2\textwidth}
    \scalebox{.8}{
      \includegraphics[width=\textwidth]{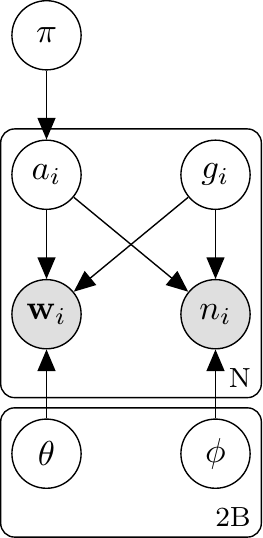}
    }
  \end{subfigure}
  \hspace{-2ex}
  \begin{subfigure}[b]{0.25\textwidth}
    \begin{small}
      \begin{tabular}{cp{.8\textwidth}} \toprule
        $a_{i}$ & Age (bin) for author $i$ \\
        $g_{i}$ & Gender of author $i$ \\
        $\vw_{i}$ & Word counts for author $i$  \\
        $n_{i}$ & First name of author $i$ \\
        $\pi$ & Prior distribution over age bins \\
        $\theta_{a,g}$ & Word distribution for age $a$ and gender $g$ \\
        $\phi_{a,g}$ & First name distribution for age $a$ and gender $g$\\
        \bottomrule
      \end{tabular}
    \end{small}
  \end{subfigure}
  \caption{Plate diagram for latent variable model of age and gender}
\label{fig:em}
\end{figure}

\paragraph{Inferring author demographics}
As described in \autoref{sssec:age-prior}, we can obtain an approximate distribution over author age and gender by linking  self-reported first names with aggregate statistics from the United States Census. To sharpen these estimates, we now consider the text as well, building a simple latent variable model in which both the name and the word counts are drawn from distributions associated with the latent age and gender~\cite{chang2010epluribus}. The model is shown in \autoref{fig:em}, and involves the following generative process:
\begin{enumerate}
  \setlength\itemsep{0pt}
  \setlength\parsep{0pt}
  \setlength\parskip{0pt}
 \item[] For each user $i \in \{1...N\}$,
 	\begin{enumerate}
  \setlength\itemsep{0pt}
  \setlength\parsep{0pt}
  \setlength\parskip{0pt}
 	\item draw the age, $a_{i} \sim \text{Categorical}(\pi)$
 	\item draw the gender, $g_{i} \sim \text{Categorical}(0.5)$
 	\item draw the author's given name, $n_{i} \sim \text{Categorical}(\phi_{a_i,g_i})$
 	\item draw the word counts, $w_{i} \sim \text{Multinomial}(\theta_{a_i,g_i})$,
	\end{enumerate}
 \end{enumerate}
where we elide the second parameter of the multinomial distribution, the total word count. We use expectation-maximization to perform inference in this model, binning the latent age variable into four groups: 0-17, 18-29, 30-39, above 40.\footnote{Binning is often employed in work on text-based age prediction~\cite{garera2009modeling,rao2010classifying,rosenthal2011age}; it enables word and name counts to be shared over multiple ages, and avoids the complexity inherent in regressing a high-dimensional textual predictors against a numerical variable.} Because the distribution of names given demographics is available from the Social Security data, we clamp the value of $\phi$ throughout the EM procedure. Other work in the domain of demographic prediction often involves more complex methods~\cite{nguyen142014gender,volkova2015bayesian}, but since it is not the focus of our research, we take a relatively simple approach here, assuming no labeled data for demographic attributes.

\subsection{Results}

\paragraph{Linguistic differences by dataset} 
We first consider the impact of the data acquisition technique on the lexical features associated with each city. The keywords identified in \gpsmsa\ dataset feature more geographically-specific non-standard words, which occur at a rate of $3.9\times 10^{-4}$ in \gpsmsa, versus $2.6\times 10^{-4}$ in \locmsa; this difference is statistically significant ($p < .05, t=3.2$).\footnote{We employ a paired t-test, comparing the difference in frequency for each word across the two datasets. Since we cannot test the complete set of entity names or non-standard words, this quantifies whether the observed difference is robust across the subset of the vocabulary that we have selected.} For entity names, the difference between datasets was not significant, with a rate of $4.0 \times 10^{-3}$ for \gpsmsa, and $3.7 \times 10^{-3}$ for \locmsa. Note that these rates include only the non-standard words and entity names detected by SAGE as among the top 25 most distinctive for one of the ten largest cities in the US; of course there are many other relevant terms that are below this threshold. 

In a pilot study of the \gpscty\ data, we found few linguistic differences from \gpsmsa, in either the aggregate word-group frequencies or the SAGE word lists --- despite the geographical imbalances shown in \autoref{tab:l1-distance} and \autoref{fig:nyc_pop}. Informal examination of specific counties shows some expected differences: for example, Clayton County, which hosts Atlanta's Hartsfield-Jackson airport, includes terms related to air travel, and other counties include mentions of local cities and business districts. But the aggregate statistics for underrepresented counties are not substantially different from those of overrepresented counties, and are largely unaffected by county-based resampling.

\begin{figure}
  \small
  \centering
  \includegraphics[width=.5\textwidth]{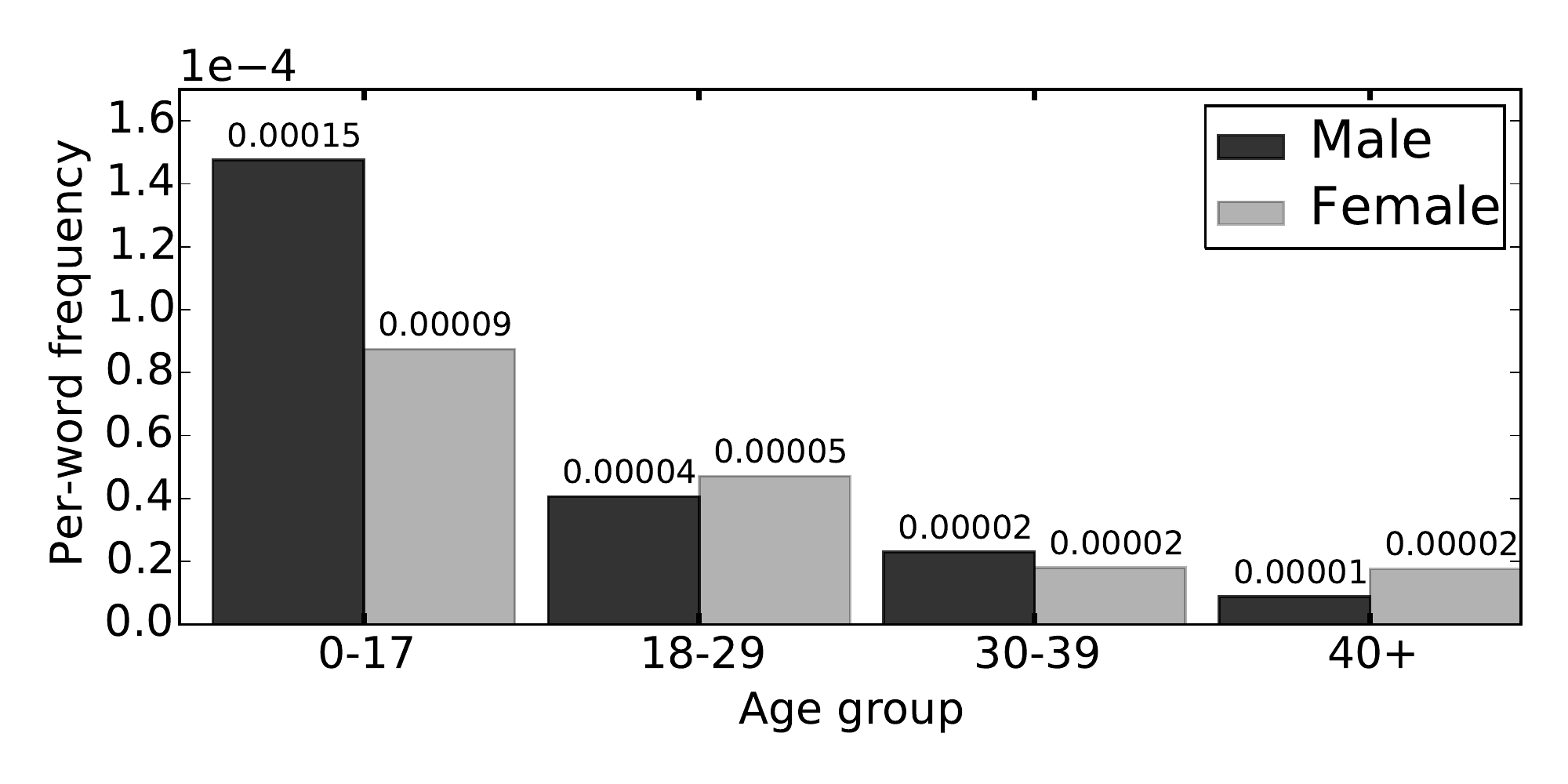}\\
  (a) non-standard words

  \includegraphics[width=.5\textwidth]{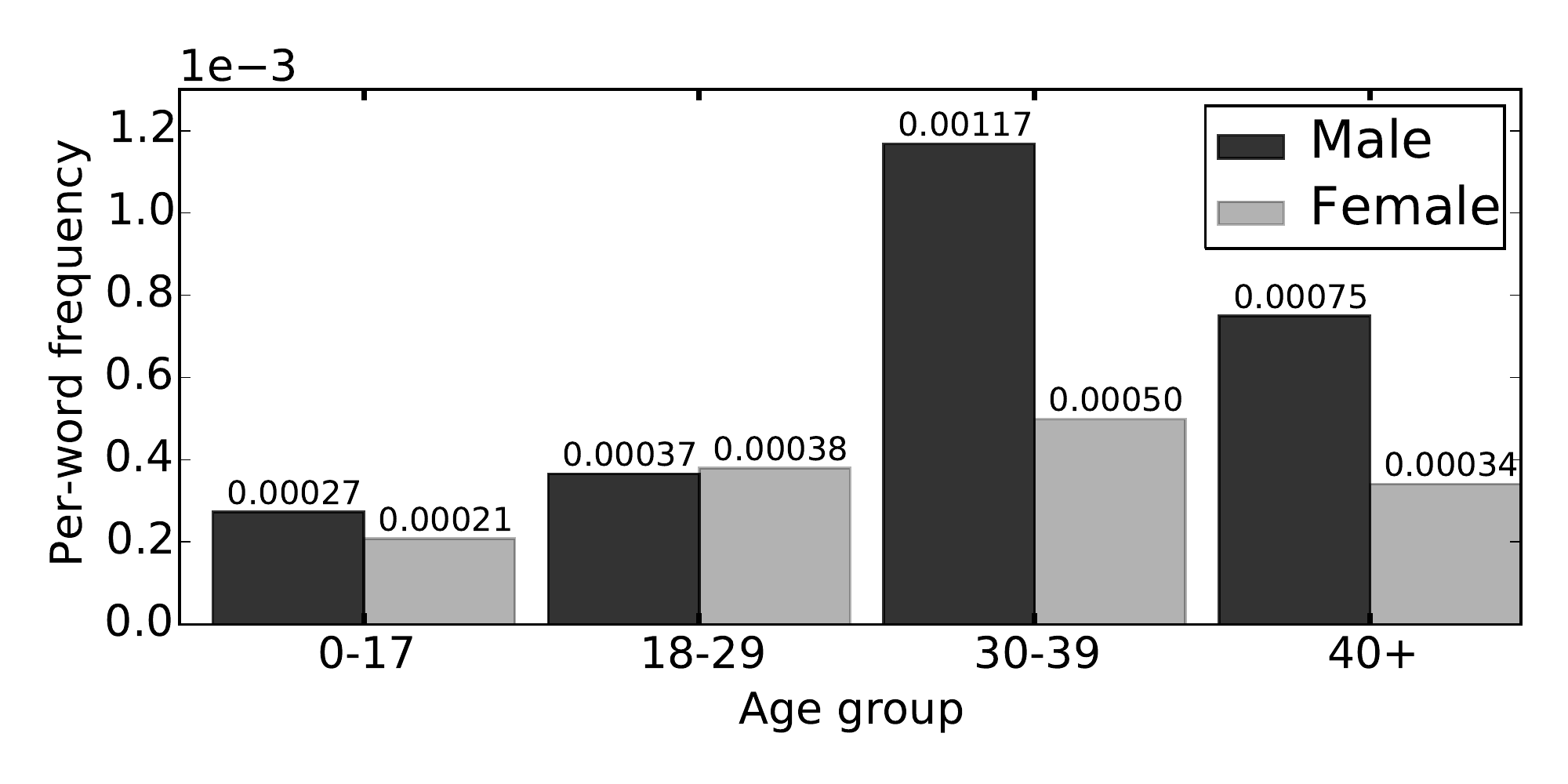}\\
  (b) entity names
  \caption{Aggregate statistics for geographically-specific non-standard words and entity names across imputed demographic groups, from the \gpsmsa\ sample.}
  \label{fig:age-gender-agg}
\end{figure}

\begin{table*}
\begin{small}
\begin{tabular}{llp{2.5in}p{2.65in}}
  \toprule
  Age & Sex & New York & Dallas \\
  \midrule
  \multirow{2}{*}{0-17} & F 
  & \example{niall, ilysm, hemmings, stalk, ily} 
  & \example{fanuary, idol, lmbo, lowkey, jonas}\\
  & M 
  & \example{ight, technique, kisses, lesbian, dicks} 
  & \example{homies, daniels, oomf, teenager, brah}\\[3pt]
  \multirow{2}{*}{18-29} & F 
  & \example{roses, castle, hmmmm, chem, sinking}
  & \example{socially, coma, hubby, bra, swimming}\\
  & M 
  & \example{drunken, manhattan, spoiler, guardians, gonna}
  & \example{harden, watt, astros, rockets, mavs}\\[3pt]
  \multirow{2}{*}{30-39} 
  & F 
  & \example{suite, nyc, colleagues, york, portugal}
  & \example{astros, sophia, recommendations, houston, prepping}\\
  & M 
  & \example{mets, effectively, cruz, founder, knicks}
  & \example{texans, rockets, embarrassment, tcu, mississippi}\\[3pt]
  \multirow{2}{*}{40+} 
  & F 
  & \example{cultural, affected, encouraged, proverb, unhappy}
  & \example{determine, islam, rejoice, psalm, responsibility}\\
  & M 
  & \example{reuters, investors, shares, lawsuit, theaters}
  & \example{mph, wazers, houston, tx, harris}\\
  \bottomrule
\end{tabular}
\end{small}
\caption{Most characteristic words for demographic subsets of each city, as compared with the overall average word distribution}
\label{tab:sage-s1-age-gender}
\end{table*}

\paragraph{Demographics}
Aggregate linguistic statistics for demographic groups are shown in \autoref{fig:age-gender-agg}. Men use significantly more geographically-specific entity names than women ($p \ll .01, t = 8.0$), but gender differences for geographically-specific non-standard words are not significant ($p \approx .2$).\footnote{But see \newcite{bamman2014gender} for a much more detailed discussion of gender and standardness.} Younger people use significantly more geographically-specific non-standard words than older people (ages 0--29 versus 30+, $p \ll .01, t=7.8$), and older people mention significantly more geographically-specific entity names ($p \ll .01, t=5.1$). Of particular interest is the intersection of age and gender: the use of geographically-specific non-standard words decreases with age much more profoundly for men than for women; conversely, the frequency of mentioning geographically-specific entity names increases dramatically with age for men, but to a much lesser extent for women. The observation that high-level patterns of geographically-oriented language are more age-dependent for men than for women suggests an intriguing site for future research on the intersectional construction of linguistic identity.

For a more detailed view, we apply SAGE to identify the most salient lexical items for each MSA, subgrouped by age and gender. \autoref{tab:sage-s1-age-gender} shows word lists for New York (the largest MSA) and Dallas (the 5th-largest MSA), using the \gpsmsa\ sample. Non-standard words tend to be used by the youngest authors: \example{ilysm} ('I love you so much'), \example{ight} ('alright'), \example{oomf} ('one of my followers'). Older authors write more about local entities (\example{manhattan, nyc, houston}), with men focusing on sports-related entities (\example{harden, watt, astros, mets, texans}), and women above the age of 40 emphasizing religiously-oriented terms (\example{proverb, islam, rejoice, psalm}).

%% file: prediction.tex
\section{Impact on text-based geolocation}
\label{sec:prediction}
A major application of geotagged social media is to predict the geolocation of individuals based on their text~\cite{eisenstein2010latent,cheng2010you,wing2011simple,hong2012discovering,han2014text}. Text-based geolocation has obvious commercial implications for location-based marketing and opinion analysis; it is also potentially useful for researchers who want to measure geographical phenomena in social media, and wish to access a larger set of individuals than those who provide their locations explicitly.

Previous research has obtained impressive accuracies for text-based geolocation: for example, \newcite{hong2012discovering} report a median error of 120~km, which is roughly the distance from Los Angeles to San Diego, in a prediction space over the entire continental United States. These accuracies are computed on test sets that were acquired through the same procedures as the training data, so if the acquisition procedures have geographic and demographic biases, then the resulting accuracy estimates will be biased too. Consequently, they may be overly optimistic (or pessimistic!) for some types of authors. In this section, we explore where these text-based geolocation methods are most and least accurate.

\subsection{Methods}
Our data is drawn from the ten largest metropolitan areas in the United States, and we formulate text-based geolocation as a ten-way classification problem, similar to~\newcite{han2014text}.\footnote{Many previous papers have attempted to identify the precise latitude and longitude coordinates of individual authors, but obtaining high accuracy on this task involves much more complex methods, such as latent variable models~\cite{eisenstein2010latent,hong2012discovering}, or multilevel grid structures~\cite{cheng2010you,roller2012supervised}. Tuning such models can be challenging, and the resulting accuracies might be affected by initial conditions or hyperparameters. We therefore focus on classification, employing the familiar and well-understood method of logistic regression.} Using our user-balanced samples, we apply ten-fold cross validation, and tune the regularization parameter on a development fold, using the vocabulary of the sample as features. 

\subsection{Results}
Many author-attribute prediction tasks become substantially easier as more data is available~\cite{burger2011discriminating}, and text-based geolocation is no exception. Since \gpsmsa\ and \locmsa\ have very different usage rates (\autoref{fig:usage}), perceived differences in accuracy may be purely attributable to the amount of data available per user, rather than to users in one group being inherently harder to classify than another. For this reason, we bin users by the number of messages in our sample of their timeline, and report results separately for each bin. All errorbars represent 95\% confidence intervals.

\begin{figure*}
\centering
\begin{subfigure}[b]{0.32\textwidth}
\includegraphics[width=\textwidth]{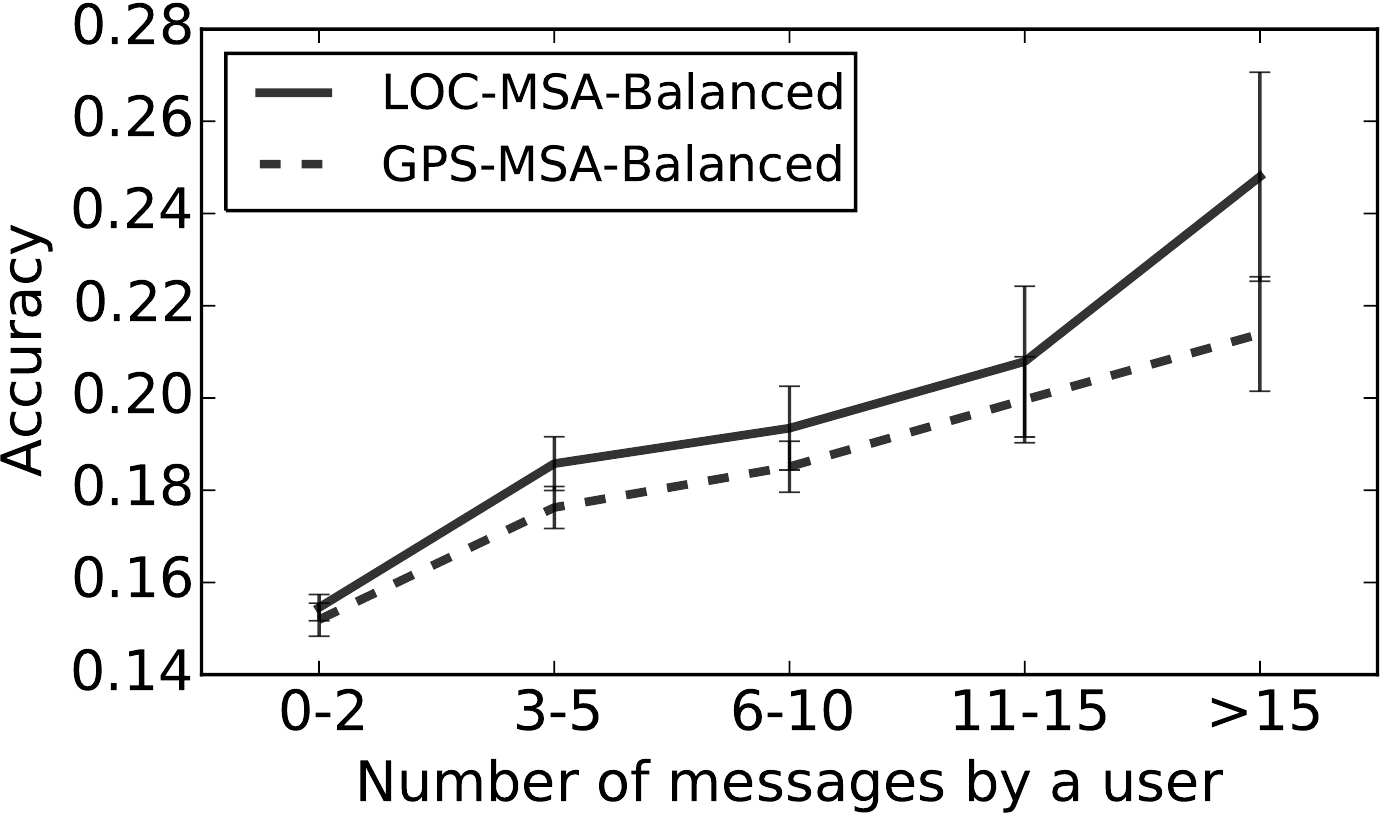}
\caption{Classification accuracy by sampling technique}
\label{fig:acc-sampling}
\end{subfigure}
~
\begin{subfigure}[b]{0.32\textwidth}
\includegraphics[width=\textwidth]{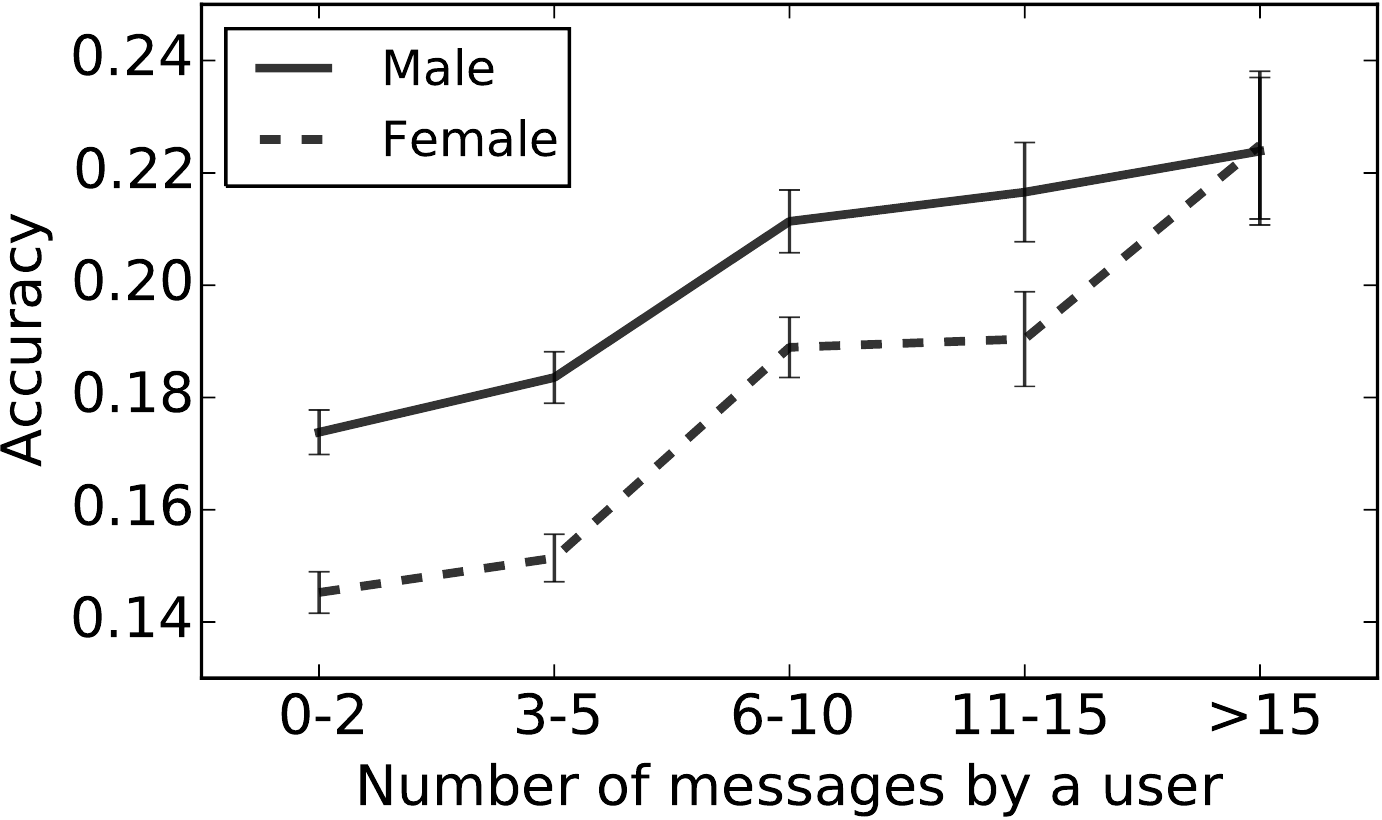}
\caption{Classification accuracy by user gender}
\label{fig:acc-gender}
\end{subfigure}
~
\begin{subfigure}[b]{0.32\textwidth}
\includegraphics[width=\textwidth]{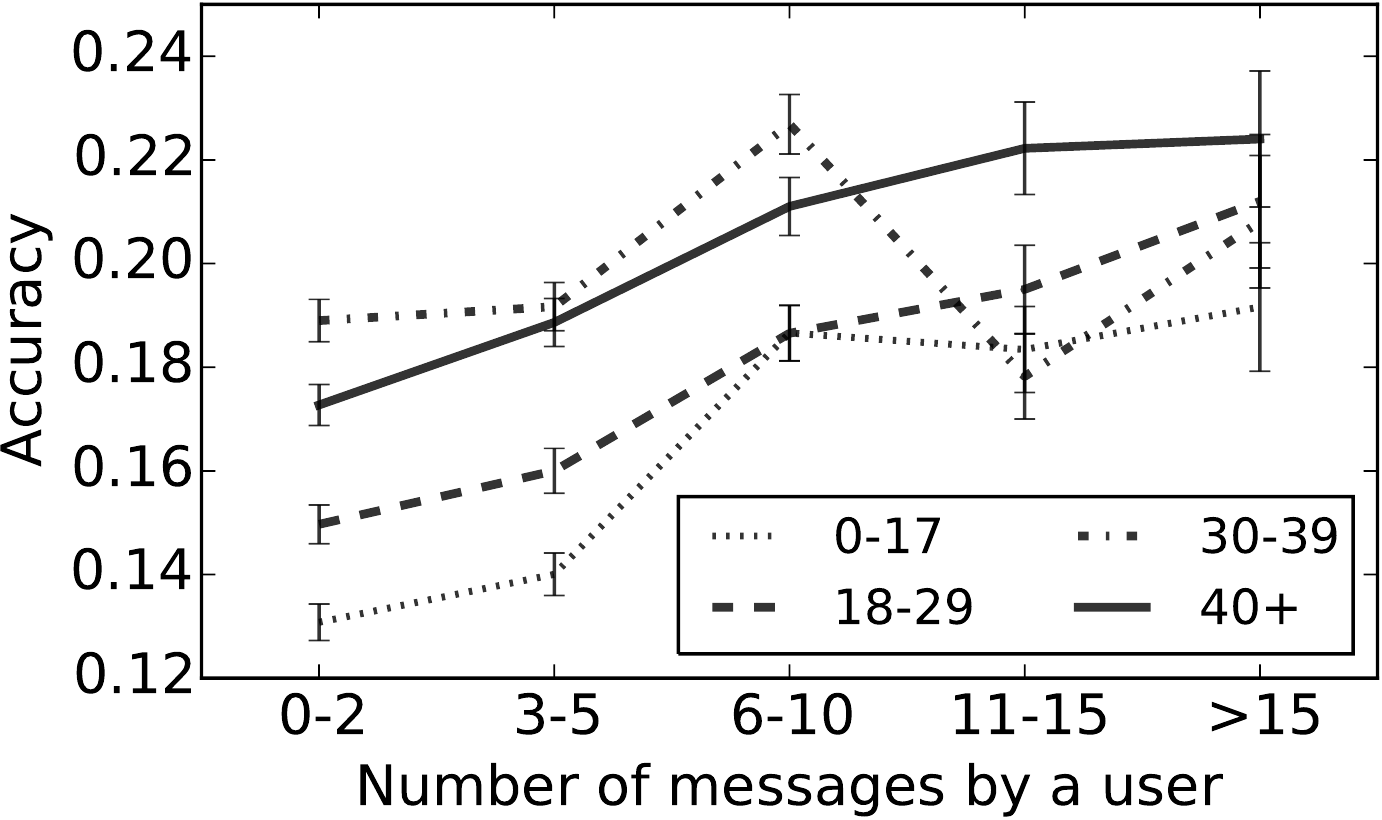}
\caption{Classification accuracy by imputed age}
\label{fig:acc-age}
\end{subfigure}
\caption{Classification accuracies}\label{fig:acc}
\end{figure*}

\paragraph{GPS versus location}

As seen in \autoref{fig:acc-sampling}, there is little difference in accuracy across sampling techniques: the location-based sample is slightly easier to geolocate at each usage bin, but the difference is not statistically significant. However, due to the higher average usage rate in \gpsmsa (see \autoref{fig:usage}), the overall accuracy for a sample of users will appear to be higher on this data.

\paragraph{Demographics}

Next, we measure classification accuracy by gender and age, using the posterior distribution from the expectation-maximization algorithm to predict the gender of each user (broadly similar results are obtained by using the prior distribution). For this experiment, we focus on the \gpsmsa\ sample. As shown in \autoref{fig:acc-gender}, text-based geolocation is consistently more accurate for male authors, across almost the entire spectrum of usage rates. As shown in \autoref{fig:acc-age}, older users also tend to be easier to geolocate: at each usage level, the highest accuracy goes to one of the two older groups, and the difference is significant in almost every case. As discussed in \autoref{sec:ling}, older male users tend to mention many entities, particularly sports-related terms; these terms are apparently more predictive than the non-standard spellings and slang favored by younger authors.



%% file: related.tex
\section{Related Work}

Several researchers have studied how adoption of Internet technology varies with factors such as socioeconomic status, age, gender, and living conditions~\cite{zillien2009digital}. \newcite{hargittai2011tweet} use a longitudinal survey methodology to compare the effects of gender, race, and topics of interest on Twitter usage among young adults. Geographic variation in Twitter adoption has been considered both internationally~\cite{kulshrestha2012geographic} and within the United States, using both the Twitter location field~\cite{mislove2011understanding} and per-message GPS coordinates~\cite{hecht2014tale}. Aggregate demographic statistics of Twitter users' geographic census blocks were computed by \newcite{oconnor2010discovering} and \newcite{eisenstein2011discovering}; \newcite{malik_bias_2015}  use census demographics in spatial error model. These papers draw similar conclusions, showing that the the distribution of geotagged tweets over the US population is not random, and that higher usage is correlated with urban areas, high income, more ethnic minorities, and more young people. However, this prior work did not consider the biases introduced by relying on geotagged messages, nor the consequences for geo-linguistic analysis.

Twitter has often been used to study the geographical distribution of linguistic information, and of particular relevance are Twitter-based studies of regional dialect differences~\cite{eisenstein2010latent,doyle2014mapping,gonccalves2014crowdsourcing,eisenstein2015geographical} and text-based geolocation~\cite{cheng2010you,hong2012discovering,han2014text}. This prior work rarely considers the impact of the demographic confounds, or of the geographical biases mentioned in \autoref{sec:biases}. Recent research shows that accuracies of core language technology tasks such as part-of-speech tagging are correlated with author demographics such as author age~\cite{hovy2015tagging}; our results on location prediction are in accord with these findings. \newcite{hovy2015demographic} show that including author demographics can improve text classification, a similar approach might improve text-based geolocation as well.

We address the question about the impact of geographical biases and demographic confounds by measuring differences between three sampling techniques, in both language use and in the accuracy of text-based geolocation. Recent unpublished work proposes reweighting Twitter data to correct biases in political analysis~\cite{choy2012us} and public health~\cite{culotta2014reducing}. Our results suggest that the linguistic differences between user-supplied profile locations and per-message geotags are more significant, and that accounting for the geographical biases among geotagged messages is not sufficient to offer a representative sample of Twitter users.

%% file: discussion.tex
\section{Discussion}
Geotagged Twitter data offers an invaluable resource for studying the interaction of language and geography, and is helping to usher in a new generation of location-aware language technology. This makes critical investigation of the nature of this data source particularly important. This paper uncovers demographic confounds in the linguistic analysis of geo-located Twitter data, but is limited to demographics that can be readily induced from given names. A key task for future work is to quantify the representativeness of geotagged Twitter data with respect to factors such as race and socioeconomic status, while holding geography constant. However, these features may be more difficult to impute from names alone. 
Another crucial task is to expand this investigation beyond the United States, as the varying patterns of use for social media across countries~\cite{pew2012social} implies that the findings here cannot be expected to generalize to every international context.

%% file: ack.tex
\paragraph{Acknowledgments}
Thanks to the anonymous reviewers for their useful and constructive feedback on our submission. The following members of the Georgia Tech Computational Linguistics Laboratory offered feedback throughout the research process: Naman Goyal, Yangfeng Ji, Vinodh Krishan, Ana Smith, Yijie Wang, and Yi Yang. This research was supported by the National Science Foundation under awards IIS-1111142 and RI-1452443, by the National Institutes of Health under award number R01GM112697-01, and by the Air Force Office of Scientific Research. The content is solely the responsibility of the authors and does not necessarily represent the official views of these sponsors.

%% file: main-arxiv.bbl
\begin{thebibliography}{}

\bibitem[\protect\citename{Bamman \bgroup et al.\egroup
  }2014]{bamman2014gender}
David Bamman, Jacob Eisenstein, and Tyler Schnoebelen.
\newblock 2014.
\newblock Gender identity and lexical variation in social media.
\newblock {\em Journal of Sociolinguistics}, 18(2):135--160.

\bibitem[\protect\citename{Broniatowski \bgroup et al.\egroup
  }2013]{broniatowski2013national}
David~A Broniatowski, Michael~J Paul, and Mark Dredze.
\newblock 2013.
\newblock National and local influenza surveillance through twitter: An
  analysis of the 2012-2013 influenza epidemic.
\newblock {\em PloS one}, 8(12):e83672.

\bibitem[\protect\citename{Burger \bgroup et al.\egroup
  }2011]{burger2011discriminating}
John~D. Burger, John Henderson, George Kim, and Guido Zarrella.
\newblock 2011.
\newblock Discriminating gender on twitter.
\newblock In {\em Proceedings of the Conference on Empirical Methods in Natural
  Language Processing}.

\bibitem[\protect\citename{Caldarelli \bgroup et al.\egroup
  }2014]{caldarelli2014multi}
Guido Caldarelli, Alessandro Chessa, Fabio Pammolli, Gabriele Pompa,
  Michelangelo Puliga, Massimo Riccaboni, and Gianni Riotta.
\newblock 2014.
\newblock {A multi-level geographical study of Italian political elections from
  Twitter Data}.
\newblock {\em PloS one}, 9(5):e95809.

\bibitem[\protect\citename{Chang \bgroup et al.\egroup
  }2010]{chang2010epluribus}
Jonathan Chang, Itamar Rosenn, Lars Backstrom, and Cameron Marlow.
\newblock 2010.
\newblock e{P}luribus: Ethnicity on social networks.
\newblock In {\em {Proceedings of the International Conference on Web and
  Social Media (ICWSM)}}, pages 18--25, Menlo Park, California. {AAAI}
  Publications.

\bibitem[\protect\citename{Cheng \bgroup et al.\egroup }2010]{cheng2010you}
Zhiyuan Cheng, James Caverlee, and Kyumin Lee.
\newblock 2010.
\newblock You are where you tweet: a content-based approach to geo-locating
  twitter users.
\newblock In {\em Proceedings of the International Conference on Information
  and Knowledge Management ({CIKM})}, pages 759--768.

\bibitem[\protect\citename{Choy \bgroup et al.\egroup }2012]{choy2012us}
Murphy Choy, Michelle Cheong, Ma~Nang Laik, and Koo~Ping Shung.
\newblock 2012.
\newblock Us presidential election 2012 prediction using census corrected
  twitter model.
\newblock {\em arXiv preprint arXiv:1211.0938}.

\bibitem[\protect\citename{Culotta}2014]{culotta2014reducing}
Aron Culotta.
\newblock 2014.
\newblock Reducing sampling bias in social media data for county health
  inference.
\newblock In {\em Joint Statistical Meetings Proceedings}.

\bibitem[\protect\citename{Doyle}2014]{doyle2014mapping}
Gabriel Doyle.
\newblock 2014.
\newblock Mapping dialectal variation by querying social media.
\newblock In {\em {Proceedings of the European Chapter of the Association for
  Computational Linguistics (EACL)}}, pages 98--106, Stroudsburg, Pennsylvania.
  Association for Computational Linguistics.

\bibitem[\protect\citename{Dredze \bgroup et al.\egroup
  }2013]{dredze2013carmen}
Mark Dredze, Michael~J Paul, Shane Bergsma, and Hieu Tran.
\newblock 2013.
\newblock Carmen: A {Twitter} geolocation system with applications to public
  health.
\newblock In {\em AAAI Workshop on Expanding the Boundaries of Health
  Informatics Using Artificial Intelligence}, pages 20--24.

\bibitem[\protect\citename{Eisenstein \bgroup et al.\egroup
  }2010]{eisenstein2010latent}
Jacob Eisenstein, Brendan O'Connor, Noah~A. Smith, and Eric~P. Xing.
\newblock 2010.
\newblock A latent variable model for geographic lexical variation.
\newblock In {\em {Proceedings of Empirical Methods for Natural Language
  Processing (EMNLP)}}, pages 1277--1287, Stroudsburg, Pennsylvania.
  Association for Computational Linguistics.

\bibitem[\protect\citename{Eisenstein \bgroup et al.\egroup
  }2011a]{eisenstein2011sparse}
Jacob Eisenstein, Amr Ahmed, and Eric~P. Xing.
\newblock 2011a.
\newblock Sparse additive generative models of text.
\newblock In {\em {Proceedings of the International Conference on Machine
  Learning (ICML)}}, pages 1041--1048, Seattle, WA.

\bibitem[\protect\citename{Eisenstein \bgroup et al.\egroup
  }2011b]{eisenstein2011discovering}
Jacob Eisenstein, Noah~A. Smith, and Eric~P. Xing.
\newblock 2011b.
\newblock Discovering sociolinguistic associations with structured sparsity.
\newblock In {\em {Proceedings of the Association for Computational Linguistics
  (ACL)}}, pages 1365--1374, Portland, OR.

\bibitem[\protect\citename{Eisenstein}2015]{eisenstein2015geographical}
Jacob Eisenstein.
\newblock 2015.
\newblock Written dialect variation in online social media.
\newblock In Charles Boberg, John Nerbonne, and Dom Watt, editors, {\em
  Handbook of Dialectology}. Wiley.

\bibitem[\protect\citename{Eleta and Golbeck}2014]{eleta2014multilingual}
Irene Eleta and Jennifer Golbeck.
\newblock 2014.
\newblock Multilingual use of twitter: Social networks at the language
  frontier.
\newblock {\em Computers in Human Behavior}, 41:424--432.

\bibitem[\protect\citename{Garera and Yarowsky}2009]{garera2009modeling}
Nikesh Garera and David Yarowsky.
\newblock 2009.
\newblock Modeling latent biographic attributes in conversational genres.
\newblock In {\em {Proceedings of the Association for Computational Linguistics
  (ACL)}}, pages 710--718, Suntec, Singapore.

\bibitem[\protect\citename{Gon{\c{c}}alves and
  S{\'a}nchez}2014]{gonccalves2014crowdsourcing}
Bruno Gon{\c{c}}alves and David S{\'a}nchez.
\newblock 2014.
\newblock Crowdsourcing dialect characterization through twitter.
\newblock {\em PloS one}, 9(11):e112074.

\bibitem[\protect\citename{Han \bgroup et al.\egroup }2014]{han2014text}
Bo~Han, Paul Cook, and Timothy Baldwin.
\newblock 2014.
\newblock Text-based twitter user geolocation prediction.
\newblock {\em Journal of Artificial Intelligence Research {(JAIR)}},
  49:451--500.

\bibitem[\protect\citename{Hargittai and Litt}2011]{hargittai2011tweet}
Eszter Hargittai and Eden Litt.
\newblock 2011.
\newblock The tweet smell of celebrity success: Explaining variation in twitter
  adoption among a diverse group of young adults.
\newblock {\em New Media \& Society}, 13(5):824--842.

\bibitem[\protect\citename{Hecht and Stephens}2014]{hecht2014tale}
Brent Hecht and Monica Stephens.
\newblock 2014.
\newblock A tale of cities: Urban biases in volunteered geographic information.
\newblock In {\em {Proceedings of the International Conference on Web and
  Social Media (ICWSM)}}, pages 197--205, Menlo Park, California. {AAAI}
  Publications.

\bibitem[\protect\citename{Hecht \bgroup et al.\egroup }2011]{hecht2011tweets}
Brent Hecht, Lichan Hong, Bongwon Suh, and Ed~H Chi.
\newblock 2011.
\newblock {Tweets from Justin Bieber's heart: the dynamics of the location
  field in user profiles}.
\newblock In {\em {Proceedings of Human Factors in Computing Systems (CHI)}},
  pages 237--246.

\bibitem[\protect\citename{Hong \bgroup et al.\egroup
  }2012]{hong2012discovering}
Liangjie Hong, Amr Ahmed, Siva Gurumurthy, Alexander~J. Smola, and Kostas
  Tsioutsiouliklis.
\newblock 2012.
\newblock Discovering geographical topics in the twitter stream.
\newblock In {\em {Proceedings of the Conference on World-Wide Web (WWW)}},
  pages 769--778, Lyon, France.

\bibitem[\protect\citename{Hovy and S{\o}gaard}2015]{hovy2015tagging}
Dirk Hovy and Anders S{\o}gaard.
\newblock 2015.
\newblock Tagging performance correlates with author age.
\newblock In {\em {Proceedings of the Association for Computational Linguistics
  (ACL)}}, pages 483--488, Beijing, China.

\bibitem[\protect\citename{Hovy}2015]{hovy2015demographic}
Dirk Hovy.
\newblock 2015.
\newblock Demographic factors improve classification performance.
\newblock In {\em {Proceedings of the Association for Computational Linguistics
  (ACL)}}, pages 752--762, Beijing, China.

\bibitem[\protect\citename{Kulshrestha \bgroup et al.\egroup
  }2012]{kulshrestha2012geographic}
Juhi Kulshrestha, Farshad Kooti, Ashkan Nikravesh, and Krishna~P. Gummadi.
\newblock 2012.
\newblock {Geographic Dissection of the Twitter Network}.
\newblock In {\em {Proceedings of the International Conference on Web and
  Social Media (ICWSM)}}, Menlo Park, California. {AAAI} Publications.

\bibitem[\protect\citename{Lenhart}2015]{lenhart2015mobile}
Amanda Lenhart.
\newblock 2015.
\newblock Mobile access shifts social media use and other online activities.
\newblock Technical report, Pew Research Center, April.

\bibitem[\protect\citename{Longley \bgroup et al.\egroup }2015]{longley2015}
P.~A. Longley, M.~Adnan, and G.~Lansley.
\newblock 2015.
\newblock The geotemporal demographics of twitter usage.
\newblock {\em Environment and Planning A}, 47(2):465--484.

\bibitem[\protect\citename{Malik \bgroup et al.\egroup }2015]{malik_bias_2015}
Momin Malik, Hemank Lamba, Constantine Nakos, and J\"{u}rgen Pfeffer.
\newblock 2015.
\newblock Population bias in geotagged tweets.
\newblock In {\em Papers from the 2015 ICWSM Workshop on Standards and
  Practices in Large-Scale Social Media Research}, pages 18--27. The {AAAI}
  Press.

\bibitem[\protect\citename{Mislove \bgroup et al.\egroup
  }2011]{mislove2011understanding}
Alan Mislove, Sune Lehmann, Yong-Yeol Ahn, Jukka-Pekka Onnela, and J.~Niels
  Rosenquist.
\newblock 2011.
\newblock Understanding the demographics of twitter users.
\newblock In {\em {Proceedings of the International Conference on Web and
  Social Media (ICWSM)}}, pages 554--557, Menlo Park, California. {AAAI}
  Publications.

\bibitem[\protect\citename{Monroe \bgroup et al.\egroup
  }2008]{monroe2008fightin}
Burt~L Monroe, Michael~P Colaresi, and Kevin~M Quinn.
\newblock 2008.
\newblock Fightin'words: Lexical feature selection and evaluation for
  identifying the content of political conflict.
\newblock {\em Political Analysis}, 16(4):372--403.

\bibitem[\protect\citename{Nguyen \bgroup et al.\egroup
  }2014]{nguyen142014gender}
Dong Nguyen, Dolf Trieschnigg, A~Seza Dogru{\"o}z, Rilana Gravel, Mari{\"e}t
  Theune, Theo Meder, and Franciska de~Jong.
\newblock 2014.
\newblock Why gender and age prediction from tweets is hard: Lessons from a
  crowdsourcing experiment.
\newblock In {\em {Proceedings of the International Conference on Computational
  Linguistics (COLING)}}, pages 1950--1961.

\bibitem[\protect\citename{O'Connor \bgroup et al.\egroup
  }2010]{oconnor2010discovering}
Brendan O'Connor, Jacob Eisenstein, Eric~P. Xing, and Noah~A. Smith.
\newblock 2010.
\newblock A mixture model of demographic lexical variation.
\newblock In {\em Proceedings of {NIPS} Workshop on Machine Learning for Social
  Computing}, Vancouver.

\bibitem[\protect\citename{{Pew Research Center}}2012]{pew2012social}
{Pew Research Center}.
\newblock 2012.
\newblock Social networking popular across globe.
\newblock Technical report, December.

\bibitem[\protect\citename{Poblete \bgroup et al.\egroup }2011]{poblete2011all}
Barbara Poblete, Ruth Garcia, Marcelo Mendoza, and Alejandro Jaimes.
\newblock 2011.
\newblock Do all birds tweet the same? characterizing {Twitter} around the
  world.
\newblock In {\em Proceedings of the International Conference on Information
  and Knowledge Management ({CIKM})}, pages 1025--1030. ACM.

\bibitem[\protect\citename{Rao \bgroup et al.\egroup }2010]{rao2010classifying}
Delip Rao, David Yarowsky, Abhishek Shreevats, and Manaswi Gupta.
\newblock 2010.
\newblock Classifying latent user attributes in twitter.
\newblock In {\em Proceedings of Workshop on Search and mining user-generated
  contents}.

\bibitem[\protect\citename{Roller \bgroup et al.\egroup
  }2012]{roller2012supervised}
Stephen Roller, Michael Speriosu, Sarat Rallapalli, Benjamin Wing, and Jason
  Baldridge.
\newblock 2012.
\newblock Supervised text-based geolocation using language models on an
  adaptive grid.
\newblock In {\em {Proceedings of Empirical Methods for Natural Language
  Processing (EMNLP)}}, pages 1500--1510.

\bibitem[\protect\citename{Rosenthal and McKeown}2011]{rosenthal2011age}
Sara Rosenthal and Kathleen McKeown.
\newblock 2011.
\newblock Age prediction in blogs: A study of style, content, and online
  behavior in pre- and {Post-Social} media generations.
\newblock In {\em {Proceedings of the Association for Computational Linguistics
  (ACL)}}, pages 763--772, Portland, OR.

\bibitem[\protect\citename{Solorio and Liu}2008]{solorio2008learning}
Thamar Solorio and Yang Liu.
\newblock 2008.
\newblock Learning to predict code-switching points.
\newblock In {\em {Proceedings of Empirical Methods for Natural Language
  Processing (EMNLP)}}, pages 973--981, Honolulu, HI, October. Association for
  Computational Linguistics.

\bibitem[\protect\citename{Volkova and Durme}2015]{volkova2015bayesian}
Svitlana Volkova and Benjamin~Van Durme.
\newblock 2015.
\newblock Online bayesian models for personal analytics in social media.
\newblock In {\em {Proceedings of the National Conference on Artificial
  Intelligence (AAAI)}}, pages 2325--2331.

\bibitem[\protect\citename{Wing and Baldridge}2011]{wing2011simple}
Benjamin Wing and Jason Baldridge.
\newblock 2011.
\newblock Simple supervised document geolocation with geodesic grids.
\newblock In {\em {Proceedings of the Association for Computational Linguistics
  (ACL)}}, pages 955--964, Portland, OR.

\bibitem[\protect\citename{Yardi \bgroup et al.\egroup
  }2009]{yardi2009detecting}
Sarita Yardi, Daniel Romero, Grant Schoenebeck, et~al.
\newblock 2009.
\newblock Detecting spam in a twitter network.
\newblock {\em First Monday}, 15(1).

\bibitem[\protect\citename{Zickuhr}2013]{zickuhr2013location}
Kathryn Zickuhr.
\newblock 2013.
\newblock Location-based services.
\newblock Technical report, Pew Research Center, Septmeber.

\bibitem[\protect\citename{Zillien and Hargittai}2009]{zillien2009digital}
Nicole Zillien and Eszter Hargittai.
\newblock 2009.
\newblock Digital distinction: Status-specific types of internet usage*.
\newblock {\em Social Science Quarterly}, 90(2):274--291.

\end{thebibliography}
